\documentclass[runningheads]{llncs}

\usepackage{graphicx}
\usepackage{multirow}	% 表格跨行
\usepackage{arydshln}   % 虚线
\usepackage{color}		% 字体颜色
\usepackage{marvosym}	% 邮件图标
\usepackage{subfigure}  % 一行摆俩图
\begin{document}

\title{
PromptCL: Improving Event Representation via Prompt Template and Contrastive Learning
}

\titlerunning{PromptCL for Event Representation}

\author{
	Yubo Feng \and
	Lishuang Li\textsuperscript{(\Letter)} \and
	Yi Xiang\and
	Xueyang Qin
}

\authorrunning{Y. Feng et al.}

\institute{
	Dalian University of Technology, Dalian, Liaoning, China \\
	\email{argmax@126.com} \\
	\email{lils@dlut.edu.cn}
}

\maketitle              % typeset the header of the contribution

\begin{abstract}
% 研究意义
The representation of events in text plays a significant role in various NLP tasks. 
% 主流研究方法：基于预训练语言模型，利用句子粒度的对比学习方法来学习事件表示
Recent research demonstrates that contrastive learning has the ability to improve event comprehension capabilities of Pre-trained Language Models (PLMs) and enhance the performance of event representation learning. 
% 前人研究存在的问题
However, the efficacy of event representation learning based on contrastive learning and PLMs is limited by the short length of event texts. 
The length of event texts differs significantly from the text length used in the pre-training of PLMs.
As a result, there is inconsistency in the distribution of text length between pre-training and event representation learning, which may undermine the learning process of event representation based on PLMs.
% 我们的解决手段
In this study, we present PromptCL, a novel framework for event representation learning that effectively elicits the capabilities of PLMs to comprehensively capture the semantics of short event texts. 
\textbf{PromptCL} utilizes a \textbf{Prompt} template borrowed from prompt learning to expand the input text during \textbf{C}ontrastive \textbf{L}earning. 
This helps in enhancing the event representation learning by providing a structured outline of the event components. 
Moreover, we propose Subject-Predicate-Object (SPO) word order and Event-oriented Masked Language Modeling (EventMLM) to train PLMs to understand the relationships between event components. 
% 我们解决的还不错（有效性）
Our experimental results demonstrate that PromptCL outperforms state-of-the-art baselines on event related tasks.
% 我们解决的还不错（印证机理）
Additionally, we conduct a thorough analysis and demonstrate that using a prompt results in improved generalization capabilities for event representations~\footnote[1]{Our code will be available at https://github.com/YuboFeng2023/PromptCL}.


\end{abstract}

\section{Introduction}
% P1 研究意义 = 事件表示学习 + 本文工作
% P1.1 事件表示学习的研究意义
Distributed event representations are a widely-used machine-readable representation of events, known to capture meaningful features relevant to various applications~\cite{chen-etal-2021-graphplan,deng-etal-2021-ontoed,rezaee-ferraro-2021-event}.
% P1.2 提出问题：从短文本中学习事件表示是比较难的
Due to event texts are too short, capturing their semantic relationships is a challenging task.  
% P1.3 事件表示学习示例
%Fig.~\ref{fig-evt-rep-process} depicts the process of event representation and demonstrates varying degrees of distance between three distinct events.
For example, despite the greater lexical overlap between ``\underline{military launch} program'' and ``\underline{military launch} missile'' their semantic similarity is limited, but ``military launch program'' and ``army starts initiative'' display a certain degree of semantic similarity, even though they share no lexical overlap.
% P1.4 事件文本简短且精炼
%The notable feature of event text lies in its conciseness and semantic complexity, thereby necessitating an event representation model that can effectively encode event text into appropriate embeddings.

%\begin{figure}[t]
%\includegraphics[width=\textwidth]{fig-evt-rep-process.png}
%\caption{
%    As the distance between two events decreases, their semantic similarity tends to increase. 
%    For instance, despite the greater lexical overlap between ``military launch program'' and ``military launch missile'' their semantic similarity is limited. 
%    In contrast, ``military launch program'' and ``army starts initiative'' display a certain degree of semantic similarity, even though they share no lexical overlap.
%} 
%\label{fig-evt-rep-process}
%\end{figure}

% P2 领域综述 = 张量神经网络 + 预训练语言模型 + 图神经网络 + 对比学习
% P2.1 张量神经网络
In previous studies~\cite{ding-etal-2019-event,ding-etal-2015-deep,ding-etal-2016-knowledge,weber-etal-2018-event}, Neural Tensor Networks (NTNs)~\cite{socher-etal-2013-recursive} have been commonly utilized to construct event representations by composing the constituent elements of an event, i.e., (subject, predicate, object).
Nevertheless, these approaches entail a substantial compositional inductive bias and are inadequate for handling events that possess additional arguments~\cite{gao-etal-2022-improving}.
% P2.2 预训练语言模型
Recent research has demonstrated the effectiveness of employing powerful PLMs, such as BERT~\cite{devlin-etal-2019-bert}, to create flexible event representations instead of using static word vector compositions~\cite{vijayaraghavan-roy-2021-lifelong,zheng-etal-2020-incorporating}. 
Nevertheless, utilizing PLMs alone to learn event representations is insufficient to capture the complicated relationships between events.
% P2.3 图神经网络
Therefore, in order to tackle the challenge of capturing complex relations between events, some researchers have proposed the use of graph neural networks in event representation learning. 
This approach has shown to yield better performance, as demonstrated in recent studies~\cite{zheng-etal-2020-heterogeneous,zheng-etal-2022-multistructure}.
But graph neural networks are often associated with high computational complexity, which can lead to significant difficulties in training the models~\cite{zhou-etal-2020-graph}.
% P2.4 对比学习
To fully capture complicated event relations and efficiently learn event representations, Gao et al. proposed SWCC~\cite{gao-etal-2022-improving}, which leverages contrastive learning~\cite{gao-etal-2021-simcse} to improve the event comprehension ability of PLMs. 
It has achieved state-of-the-art results on event similarity tasks and downstream tasks.

% P3 提出问题 = 预训练语言模型的文本理解能力没有被充分挖掘 + 通过提示学习可以对此进行挖掘 + 为什么前人方法没有很好地挖掘
In our work, we argue that there is a rich amount of event comprehension ability in PLMs, but previous works did not make fully use of such abilities. 
%Based on existing works on prompt learning~\cite{schick-schutze-2021-exploiting,schick-and-schutze-2021-small}, we have realized that providing task descriptions to PLMs can help to elicit the knowledge contained within them for event representation learning. 
Inspired by the advancements in prompt learning~\cite{schick-schutze-2021-exploiting,schick-and-schutze-2021-small}, we have realized that providing task descriptions to PLMs can help to elicit the knowledge embedded within them. 
And then, this knowledge can be utilized to enhance event representation learning.
To learn event representations, previous works~\cite{gao-etal-2022-improving,zheng-etal-2020-incorporating} leverage contrastive learning to improve the event comprehension capacity of PLMs. 
However, they share three common limitations. 
% P3.1 事件文本较短
% 首先，事件文本的长度较短，与语言模型预训练阶段所使用文本的长度差异较大。
% BERT等主流的语言模型在预训练阶段，使用句子级语料学习较长序列的特征。
% 事件文本过短，可能导致事件表示模型无法充分地利用预训练语言模型的文本理解能力。
Firstly, the length of event text is relatively short, which differs significantly from the text length used in the pre-training of language models. 
As a result, the distribution of text length between pre-training and event representation learning is inconsistent.
This inconsistency may undermine the learning process of event representation based on PLMs.
% P3.2 PSO语序
% 其次, 事件文本采用PSO语序，与语言模型预训练阶段所使用文本的语序差异较大。
% SWCC等主流的事件表示模型采用predicate-subject-object语序来表达事件。
% 在该语序下，事件的主语和谓语被倒置。
% BERT等语言模型在预训练阶段，使用自然语言语序来表达输入文本。
% 输入文本的语序不同，可能导致事件表示模型无法充分挖掘预训练语言模型的文本理解能力。
Secondly, the Predicate-Subject-Object (PSO) word order, which is adopted by PLMs-based event representation models~\cite{gao-etal-2022-improving,zheng-etal-2020-incorporating}, is significantly different from the natural language word order used during the pre-training~\cite{devlin-etal-2019-bert}. 
In PSO word order, the inversion of subject and predicate can potentially undermine the performance, as the pre-trained MLM knowledge may have a counter-productive effect.
Because MLM in pre-training predicts a masked token based on its context~\cite{devlin-etal-2019-bert}, a change in word order can also cause the position of the context to change.
Therefore, the pre-trained MLM knowledge may be a burden for event representation learning in PSO word order.
% P3.3 MLM
Finally, the state-of-the-art model utilizes MLM loss to prevent the forgetting of token-level knowledge during the training of event representation in the PLM~\cite{gao-etal-2022-improving}.
However, the model only randomly masks one sub-word, which may not provide sufficient understanding of complex event texts~\cite{wettig-etal-2023-mask}.
%Moreover, recent studies have demonstrated that increasing the masking rate can enhance the performance of downstream tasks~\cite{wettig-etal-2023-mask}.

% P4 解决问题 = 深入挖掘预训练语言模型的文本理解能力 = SPO模版 + SPO语序 + Event MLM
We are motivated to address the above issues with the goal of eliciting the event comprehension capabilities from PLMs. 
To this end, we present \textbf{PromptCL}: a \textbf{Prompt} template-based \textbf{C}ontrastive \textbf{L}earning framework for event representation learning. 
%PromptCL deeply elicits the event comprehension abilities of PLMs by incorporating prompt template~\cite{schick-schutze-2021-exploiting,schick-and-schutze-2021-small} into event representation learning. 
%Additionally, the Subject-Predicate-Object (SPO) word order is employed to construct event components, and Event-oriented Masked Language Modeling (EventMLM) is utilized to learn the relationships between event components.
% P4.1 SPO模版
To address the first issue, we propose a novel prompt template-based contrastive learning method. 
In this approach, we incorporate a prompt template borrowed from prompt learning into contrastive learning, which comprises a description outlining the event components.
The injection of the prompt template serves two purposes: it extends the length of event texts and provides semantic guidance to PLMs. 
% P4.2 SPO语序
%To address the second issue, we propose using the Subject-Predicate-Object (SPO) word order to mitigate the problem of subject-predicate inversion. 
%The SPO word order aligns with the natural language word order.
To address the second issue, we propose using the SPO word order to solve the problem of subject-predicate inversion, which aligns with the natural language word order.
% P4.3 Event MLM
To address the final issue, we present an approach called EventMLM, which focuses on the structure of events and aims to increase the masking rate.
EventMLM not only masks entire words but also masks the complete subject, predicate, or object of the event. 
This approach trains PLMs to understand the relationships between event components.
% P4.4 总结本文贡献
Overall, our study makes the following noteworthy contributions:
\begin{itemize}
\item[$\bullet$]
We propose \textbf{PromptCL}, a simple and effective framework that improves event representation learning using PLMs. 
To the best of our knowledge, this is the first study that utilizes prompt learning and contrastive learning to elicit event representation abilities from PLMs.
\item[$\bullet$] 
We introduce prompt template-based contrastive learning that extends the length of event texts and provides semantic guidance to PLMs. 
Additionally, we introduce the SPO word order and the EventMLM method, which are designed to train PLMs to comprehend the relationships between event components.
\item[$\bullet$]
Our experimental results demonstrate that our framework outperforms previous state-of-the-art methods on event-related tasks. 
We conduct a thorough analysis of the proposed methods and demonstrate that they generate similarity scores that are more closely aligned with the ground truth labels.
\end{itemize}

\section{The Proposed Approach}

\begin{figure}[t]
\includegraphics[width=\textwidth]{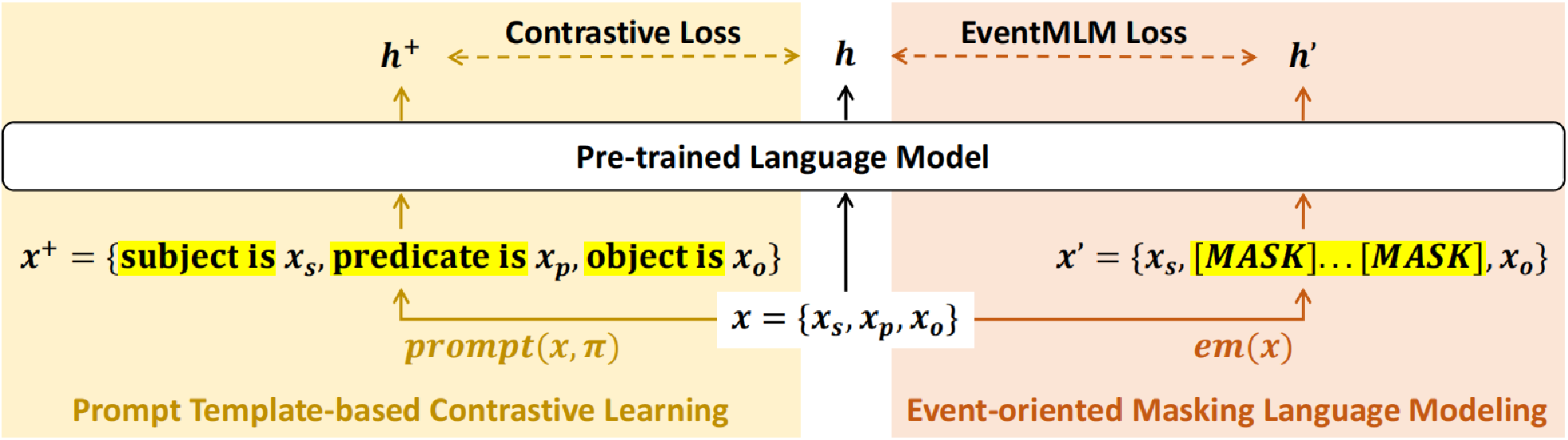}
\caption{
Architecture of PromptCL.
%    The proposed framework consists of two main components: EventMLM and prompt template-based contrastive learning. 
%    The left part of the architecture utilizes an event-oriented masking function, denoted as $em(\cdot)$, to obtain a masked event $x'$ from the input event $x = \{x_s, x_p, x_o\}$. 
%    The right part of the architecture uses a prompt template function, denoted as $prompt(x, \pi)$, to obtain a prompt-augmented event $x^+$ from $x$ with a prompt template inserting probability $\pi$. 
%    Through the simultaneous calculation of EventMLM Loss and Contrastive Loss, our objective is to enrich the event representations by deeply exploiting the text comprehension abilities of PLMs.
}
\label{fig-model}
\end{figure}

This section details our proposed approach that aims to enhance event representations by eliciting the event comprehension capabilities of PLMs.
Our approach is illustrated in Fig~\ref{fig-model}, comprising three parts: the prompt template-based contrastive learning (left), and the SPO word order (middle), and the EventMLM (right). 
%Each of these methods will be presented in detail in the following sections.

\subsection{Prompt Template-based Contrastive Learning}

The proposed contrastive learning method based on prompt templates involves augmenting an event text by randomly inserting a template using a Bernoulli distribution. 
This augmentation results in a modified event that serves as a positive example during the training of contrastive learning.

\textbf{Prompt template.} 
%The prompt template is inserted randomly according to a Bernoulli distribution. 
%Specifically, given an event $x=\{x_s, x_p, x_o\}$, the function $prompt(\cdot)$ inserts a template into the event with a probability $\pi$ following a Bernoulli distribution: 
Given an event $x=\{x_s, x_p, x_o\}$, the function $prompt(\cdot)$ inserts a template into the event with a probability $\pi$ following a Bernoulli distribution: 
\begin{equation}
x^+ = prompt(x, \pi)
\end{equation}

\noindent
and the resulting prompt-augmented event is denoted as $x^+$:
\begin{equation}
x^+ = \{subject\ is\ \underline{x_s}, predicate\ is\ \underline{x_p}, object\ is\ \underline{x_o}\}
\end{equation}

\noindent
For example, if $x=\{military, launch, missile\}$, the augmented event $x^+$ could be :
\begin{equation}
x^+=\{subject\ is\ \underline{military}, predicate\ is\ \underline{launch}, object\ is\ \underline{missile}\}	
\end{equation}

\noindent
The random insertion of the template ensures that the model is trained on a slightly diverse set of events, improving its ability to capture the core semantic meaning of events.

\textbf{Dual positive contrastive learning.} 
To augment the impact of prompt templates and enhance the diversity of positive examples, we introduce an additional positive sample, whereby an input event $x_i$ is compared not only with its prompt-augmented text $x^+_{i,1} = prompt(x_i, \pi)$, but also with another prompt-augmented text $x^+_{i,2} = prompt(x_i, \pi)$.
Drawing inspiration from~\cite{khosla-etal-2020-supervised} and based on in-batch negatives~\cite{gao-etal-2021-simcse}, we extend InfoNCE objective~\cite{van-den-oord-etal-2019-representation} to:

\begin{equation}
\mathcal{L}_{CL}=
\sum_{\alpha=\{h^+_{i,1},h^+_{i,2}\}}-\log{\frac{g(h_i, \alpha)}{g(h_i, \alpha) + \sum_{k\in{\mathcal{N}(i)}}g(h_i, h_k)}}
\end{equation}

\noindent
where $h^+_{i,1}$ and $h^+_{i,2}$ correspond to event representations of $x^+_{i,1}$ and $x^+_{i,2}$ respectively.
$k\in{\mathcal{N}(i)}$ is the index of in-batch negatives and $g(\cdot)$ is a function: $g(h_i, h_k) = \exp(h^{\top}_i h_k / \tau)$, 
where $\tau \in R^+$ is temperature.

\subsection{Subject-Predicate-Object Word Order}
 
%Unlike prior studies~\cite{zheng-etal-2020-incorporating,gao-etal-2022-improving}, we adopt a SPO word order to construct the input event.
Unlike prior studies~\cite{gao-etal-2022-improving,zheng-etal-2020-incorporating}, where PSO word orders were used to construct the input events, we use the Subject-Predicate-Object (SPO) word order in our study. 
The event text $x$ consists of three components, namely the subject $x_s$, predicate $x_p$, and object $x_o$. 
Specifically, we utilize the PLM to process an event text that consists of a sequence of tokens, following the input format represented below: 
\begin{equation}
{\rm [CLS]}\ x_s\ x_p\ x_o\ {\rm [SEP]}	
\end{equation}

\noindent
Let $s = [s_0, s_1, ... , s_L]$ be an input sequence, where $s_0$ corresponds to the $\rm [CLS]$ token and $s_L$ corresponds to the $\rm [SEP]$ token.
When given an event text as input, a PLM generates a sequence of contextualized vectors:
\begin{equation}
[v_{\rm [CLS]}, v_{x_1} , ... , v_{x_L} ] = {\rm PTM}(x)	
\end{equation}

\noindent
The representation of the $\rm [CLS]$ token, denoted by $v_{\rm [CLS]}$, serves as the first input to downstream tasks in many PLMs. 
Typically, the final representation of an input sequence is obtained by taking the $\rm [CLS]$ representation as the output, that is, $h = v_{\rm [CLS]}$.

\subsection{Event-oriented Masking Language Modeling}

To fully utilize the text comprehension ability of PLMs, we present a novel event-oriented masking function, denoted by $em(\cdot)$, which randomly masks a component of the input event using a uniform distribution. 
%The PLM is then predict the masked tokens. 
For a given event $x=\{x_s, x_p, x_o\}$, the resulting masked event is denoted as $x'$:
\begin{equation}
x' = em(x)	
\end{equation}

For example, if the predicate $x_p$ is randomly selected to be masked using a uniform distribution, we replace it with special tokens $\rm [MASK]$. 
Note that multiple tokens may be replaced by the $\rm [MASK]$ tokens.
\begin{equation}
x'=\{x_s, {\rm [MASK]...[MASK]}, x_o\}	
\end{equation}

Distinctively, our EventMLM method differs from previous work~\cite{gao-etal-2022-improving}, which merely masks a single token. 
Our proposed method not only masks several tokens but also considers the components of the event. 
Moreover, our method focuses on the event structure and trains the PLM to comprehend the relationships between the components, thus enhancing the event representation. 
In this example, the PLM needs to accurately predict the masked tokens (predicate $x_p$) by understanding the semantic relationship between the subject $x_s$ and object $x_o$.

\subsection{Model Training}

%Our approach learns event representations by simultaneously performing prompt template-based contrastive learning and EventMLM. 
The overall training objective comprises three terms:
\begin{equation}
\mathcal{L}_{overall} = \mathcal{L}_{CL} + \mathcal{L}_{EventMLM} + \mathcal{L}_{CP}	
\end{equation}

\noindent
Firstly, we have the prompt template-based contrastive learning loss ($\mathcal{L}_{CL}$), which effectively incorporates prompt templates into event representation learning. 
Secondly, the EventMLM loss ($\mathcal{L}_{EventMLM}$) aims to improve the text comprehension ability of PLMs and teaches the model to comprehend the relationships between the components of input events. 
Finally, we introduce the prototype-based clustering objective ($\mathcal{L}_{CP}$) as an auxiliary loss to cluster the events while enforcing consistency between cluster assignments produced for different augmented representations of the input event~\cite{gao-etal-2022-improving}.

\section{Experiments}

Consistent with conventional practices in event representation learning~\cite{ding-etal-2019-event,gao-etal-2022-improving,weber-etal-2018-event,zheng-etal-2020-incorporating}, we conduct an analysis of the event representations acquired through our approach on two event similarity tasks and one transfer task.

\subsection{Dataset and Implementation Details}

%To train our event representation learning models, we utilized the event triples released by Gao et al.~\cite{gao-etal-2022-improving}\footnote[1]{https://github.com/gaojun4ever/SWCC4Event}, which comprises 4,029,877 unique events.
%For the transfer task, we employed the MCNC dataset that was previously utilized by Lee and Goldwasser~\cite{lee-goldwasser-2019-multi}\footnote[2]{https://github.com/doug919/multi\_relational\_script\_learning}.
For the event representation learning models training and event similarity tasks, we utilize the datasets released by Gao et al.~\cite{gao-etal-2022-improving}\footnote[1]{https://github.com/gaojun4ever/SWCC4Event}.
For the transfer task, we use the MCNC dataset that was previously employed by Lee and Goldwasser~\cite{lee-goldwasser-2019-multi}\footnote[2]{https://github.com/doug919/multi\_relational\_script\_learning}.
It is noteworthy that above datasets explicitly specify the components of the event, indicating that they support the arbitrary organization of word order.

%We implemented our event representation model using the Texar-PyTorch package~\cite{hu-etal-2019-texar}. 
Our model begins with the checkpoint of BERT-based-uncased~\cite{devlin-etal-2019-bert}, and we utilize the $\rm [CLS]$ token representation as the event representation. 
During training, we employe an Adam optimizer with a batch size of 256. 
%The learning rate for the event representation model was set to 2e-7, while that for the prototype memory was set to 2e-5. 
The learning rate for the event representation model is set to 2e-7. 
%In our experimental setup, we utilized 10 prototypes, and the value of temperature was set to $\tau=0.3$. 
The value of temperature is set to $\tau=0.3$. 
Furthermore, we select the probability of prompt template insertion to be $\pi=0.2$.

\subsection{Event Similarity Tasks}

%Evaluating vector representations through similarity tasks is a conventional approach. 
%Weber et al.~\cite{weber-etal-2018-event} introduced two event-related similarity tasks: the Hard Similarity Task and the Transitive Sentence Similarity.

\textbf{Hard Similarity Task.} 
The objective of the hard similarity task is to assess the ability of the event representation model to differentiate between similar and dissimilar events. 
Weber et al.~\cite{weber-etal-2018-event} created a dataset (referred to as "Original") comprising two types of event pairs: one with events that have low lexical overlap but should be similar, and the other with events that have high overlap but should be dissimilar. 
The dataset consists of 230 event pairs. 
Ding et al.~\cite{ding-etal-2019-event} subsequently expanded this dataset to 1,000 event pairs (denoted as "Extended"). 
We evaluate the performance of our model on this task using Accuracy$(\%)$ as the metric, which measures the percentage of instances where the model assigns a higher cosine similarity score to the similar event pair compared to the dissimilar one.

\textbf{Transitive Sentence Similarity.} 
This dataset~\cite{kartsaklis-and-sadrzadeh-2014-study} (denoted as "Transitive") is comprised of 108 pairs of transitive sentences containing a singular subject, object, and verb (e.g., ``military launch missile''). 
Each pair is annotated with a similarity score ranging from 1 to 7, with higher scores indicating greater similarity between the two events. 
To evaluate the performance of the models, we employ Spearman’s correlation$(\rho)$ to measure the relationship between the predicted cosine similarity and the manually annotated similarity score, consistent with prior work in the field~\cite{ding-etal-2019-event,weber-etal-2018-event,zheng-etal-2020-incorporating}.

\textbf{Comparison methods.}
In our study, we conduct a comparative analysis of our proposed approach with various baseline methods. 
We group these methods into four distinct categories:

(1) Neural Tensor Networks: 
The models, \textbf{Event-comp}~\cite{weber-etal-2018-event}, \textbf{Role-factor Tensor}~\cite{weber-etal-2018-event}, \textbf{Predicate Tensor}~\cite{weber-etal-2018-event}, and \textbf{NTNIntSent}~\cite{ding-etal-2019-event}, employ Neural Tensor Networks to learn event representations.
(2) Pre-trained Language Model: 
Two event representation learning frameworks that leverage PLMs are \textbf{UniFAS}~\cite{zheng-etal-2020-incorporating} and \textbf{UniFA-S}~\cite{zheng-etal-2020-incorporating}.
%\textbf{EventBERT+DR-EMR}~\cite{zheng-etal-2020-incorporating} is a novel approach to event representation learning that integrates social commonsense knowledge with the latest developments in lifelong language learning.
(3) Graph Neural Network: 
The utilization of graph neural networks for event representation learning is employed by \textbf{HeterEvent}~\cite{zheng-etal-2020-heterogeneous} and \textbf{MulCL}~\cite{zheng-etal-2022-multistructure}.
(4) Contrastive Learning: 
\textbf{SWCC}~\cite{gao-etal-2022-improving} is a state-of-the-art framework that is based on a PLM and combines contrastive learning and clustering.

\begin{table}[t]
\caption{
	Evaluation performance on the similarity tasks. 
	The Hard Similarity Task is represented by the Original and Extended datasets.
	The Transitive Sentence Similarity is evaluated using the Transitive dataset.
}
\label{table-similarity}
\centering
\begin{tabular}{|l|l|l|l|}
	\hline
	Model & Original$(\%)$ & Extended$(\%)$ & Transitive$(\rho)$ \\
	\hline
    Event-comp~\cite{weber-etal-2018-event}            & 33.9  & 18.7  & 0.57 \\
    Predicate Tensor~\cite{weber-etal-2018-event}      & 41.0  & 25.6  & 0.63 \\
    Role-factor Tensor~\cite{weber-etal-2018-event}    & 43.5  & 20.7  & 0.64  \\
    NTN-IntSent~\cite{ding-etal-2019-event}            & 77.4  & 62.8  & 0.74 \\
	\hdashline
    UniFA~\cite{zheng-etal-2020-incorporating}         & 75.8  & 61.2  & 0.71 \\
    UniFA-S~\cite{zheng-etal-2020-incorporating}       & 78.3  & 64.1  & 0.75 \\
%	EventBERT+DR-EMR~\cite{zheng-etal-2020-incorporating}	& --  & 71.2  & --  \\
    \hdashline
%    HeterEvent$_{[W]}$~\cite{zheng-etal-2020-heterogeneous}     & 76.1  & 61.3  & 0.70 \\
%    HeterEvent$_{[E]}$~\cite{zheng-etal-2020-heterogeneous}     & 76.8  & 62.0  & 0.72 \\
    HeterEvent$_{[W+E]}$~\cite{zheng-etal-2020-heterogeneous}   & 76.6  & 62.3  & 0.73 \\
%    MulCL$_{[W]}$~\cite{zheng-etal-2022-multistructure}         & 77.7  & 63.1  & 0.73 \\
%    MulCL$_{[E]}$~\cite{zheng-etal-2022-multistructure}         & 78.5  & 63.9  & 0.75 \\
    MulCL$_{[W+E]}$~\cite{zheng-etal-2022-multistructure}       & 78.3  & 64.3  & 0.76 \\
    \hdashline
    SWCC~\cite{gao-etal-2022-improving}		& 80.9  & 72.1  & 0.82 \\
%	MPCL(Ours)                				& \textbf{81.7}  & \textbf{77.9}  & 0.81 \\    
    PromptCL(Ours)							& \textbf{81.7}  & \textbf{78.7}  & \textbf{0.82} \\
\hline
\end{tabular}
\end{table}

\textbf{Results.} 
Table~\ref{table-similarity} presents the results of various methods on the challenging similarity tasks, including hard similarity and transitive sentence similarity. 
The findings reveal that the proposed PromptCL outperforms other approaches in terms of performance. 
% PLMs
Compared to the UniFA-S approach that simply utilizes PLMs, PromptCL exhibits superior performance due to its innovative features such as prompt template and contrastive learning that better explore the text comprehension ability. 
% 具体提升效果
%Specifically, on the Original and Extended datasets, PromptCL achieves accuracy improvements of 3.4\% and 16.8\%, respectively, while Spearman's correlation on the Transitive dataset shows an improvement of 0.08.
% 对比学习
PromptCL outperforms state-of-the-art event representation methods, such as SWCC, that leverage a PLM and contrastive learning. 
% 具体提升效果
%Specifically, PromptCL achieved accuracy improvements of 0.8\% and 8.2\% on the respective datasets, while Spearman's correlation on the Transitive dataset improved by 0.01. 
% 原因
The observed enhancements can be attributed to PromptCL's thorough exploration of PLM's text comprehension capabilities via its prompt template-based contrastive learning and EventMLM techniques.
% 小结
This finding emphasizes the limited exploration of text comprehension ability in prior research and underscores the efficacy of our proposed framework, PromptCL.

\subsection{Transfer Task}

%1. 任务介绍
We conduct an evaluation of the generalization ability of event representations on the Multiple Choice Narrative Cloze (MCNC) task, which involves selecting the next event from a small set of randomly drawn events, given a sequence of events. 
%2. 与前人一致
We adopt the zero-shot transfer setting to ensure comparability with prior research~\cite{gao-etal-2022-improving}.
%To ensure comparability with prior research, we employ same unsupervised methods to assess the performance of our proposed techniques~\cite{gao-etal-2022-improving}:
%
%(1) \textbf{Random} selects a candidate at random uniformly; 
%%(2) \textbf{PPMI}~\cite{chambers-and-jurafsky-2008-unsupervised} utilizes co-occurrence information and calculates Positive PMI for event pairs; 
%(2) \textbf{PPMI} utilizes co-occurrence information and calculates Positive PMI for event pairs; 
%%(3) \textbf{BiGram}~\cite{jans-etal-2012-skip} computes bi-gram conditional probabilities based on event term frequencies; and 
%(3) \textbf{BiGram} computes bi-gram conditional probabilities based on event term frequencies; and 
%%(4) \textbf{Word2vec}~\cite{mikolov-etal-2013-efficient} leverages the word embeddings trained by the skip-gram algorithm and represents events as the sum of word embeddings of predicates and arguments. 
%(4) \textbf{Word2vec} represents events as the sum of word embeddings of predicates and arguments. 
%%We did not compare our results with supervised methods~\cite{bai-etal-2021-integrating,zhou-etal-2021-modeling,lv-etal-2020-integrating} since unsupervised ones are more suitable for evaluating the performance of event representations in isolation.

\begin{table}[t]
\caption{
Evaluation performance on the MCNC task. 
*: results reported in Gao, et al. ~\cite{gao-etal-2022-improving}
}
\label{table-mcnc-brief}
\centering
\begin{tabular}{|l|c|c|}
	\hline
    Model                           			& Accuracy(\%)    \\
    \hline
    Random   								& 20.00   \\
    PPMI*     								& 30.52   \\
    BiGram*   								& 29.67   \\
    Word2vec* 								& 37.39   \\
    \hdashline
    SWCC~\cite{gao-etal-2022-improving} 		& 44.50   \\
%	MPCL（本文工作）                			& \textbf{47.92}  \\
	PromptCL(Ours)                			& \textbf{47.06}  \\
	\hline
\end{tabular}
\end{table}

\textbf{Results.} 
The performance of various methods on the MCNC task is reported in Table~\ref{table-mcnc-brief}. 
The table indicates that the PromptCL method exhibits the highest accuracy on the MCNC task in the unsupervised setting. 
This result suggests that PromptCL has superior generalizability to downstream tasks compared to other methods in the study. 
We believe that the use of a prompt template can enhance the generalization capabilities of event representation models, as discussed in the section ``Content of prompt''.

\section{Analysis}

%In this section, we conduct a detailed analysis of the prompt templates method.

\begin{table}[t]
\caption{
	Ablation study for several methods evaluated on the similarity tasks.
	*: degenerate to PSO word order.
}
\label{table-ablation}
\centering
\begin{tabular}{|l|l|l|l|}
\hline
	Model               				& Original$(\%)$ 						& Extended$(\%)$ 						& Transitive$(\rho)$	\\
	\hline
	PromptCL            				& 81.7							& 78.7							& 0.82       					\\
	\quad w/o Prompt Template 		& 80.0(\textcolor{red}{-1.7})   & 70.8(\textcolor{red}{-7.9})   & 0.81(\textcolor{red}{-0.01}) 	\\
	\quad w/o SPO Word Order*  		& 79.9(\textcolor{red}{-1.8})   & 74.1(\textcolor{red}{-4.6})   & 0.80(\textcolor{red}{-0.02}) 	\\
	\quad w/o EventMLM    			& 80.9(\textcolor{red}{-0.8})   & 76.7(\textcolor{red}{-2.0})   & 0.76(\textcolor{red}{-0.06})	\\
	\hdashline
	BERT(InfoNCE)					& 72.1 							& 63.4 							& 0.75							\\
\hline
\end{tabular}
\end{table}

\textbf{Ablation study.}
% 导言
To evaluate the effectiveness of each component in the proposed approach, we conduct an ablation study as presented in Table~\ref{table-ablation}. 
%Specifically, we remove one component at a time and analyze the resulting performance of PromptCL on the similarity tasks.
% 第一步
We begin by investigating the impact of the prompt template method by setting the probability of inserting templates to zero. 
%Our findings indicate that this component plays a crucial role in enhancing the model's performance on both hard similarity tasks, with a more substantial impact observed on the extended hard similarity task. 
Removing the prompt template component resulted in a significant drop of 7.9 points in the model's performance on the extended hard similarity task.
% 第二步
Furthermore, we examine the effect of the SPO word order method on the similarity tasks.
%and our results suggest that this component also has a strong impact on both hard similarity tasks, but particularly the original hard similarity task. 
Removing this component led to a drop of 1.8 points in the model's performance on the original hard similarity task.
% 第三步
We also study the impact of the EventMLM method. 
%As shown in Table 2 the event component-level MLM objective improves the performance on the transitive sentence similarity task modestly.
Removing this component causes a 0.06 (maximum) point drop in performance on the transitive sentence similarity task. 
% BERT(InfoNCE)
The BERT (InfoNCE) is trained using the InfoNCE objective only.
%The BERT (InfoNCE) is trained using the InfoNCE objective only, without prompt template, dual positive contrastive learning and EventMLM.

%\begin{figure}
%\centering
%\includegraphics[width=0.45\textwidth]{fig-prompt-prob.png}
%\caption{
%	Effect of varying the probability of prompt insertion during the training phase.
%} 
%\label{fig-prompt-prob}
%\end{figure}

\begin{figure}[]
	\begin{minipage}[t]{0.5\textwidth}	%并排放两张图片，每张占页面的0.5
	\centering
		\includegraphics[width=0.96\textwidth]{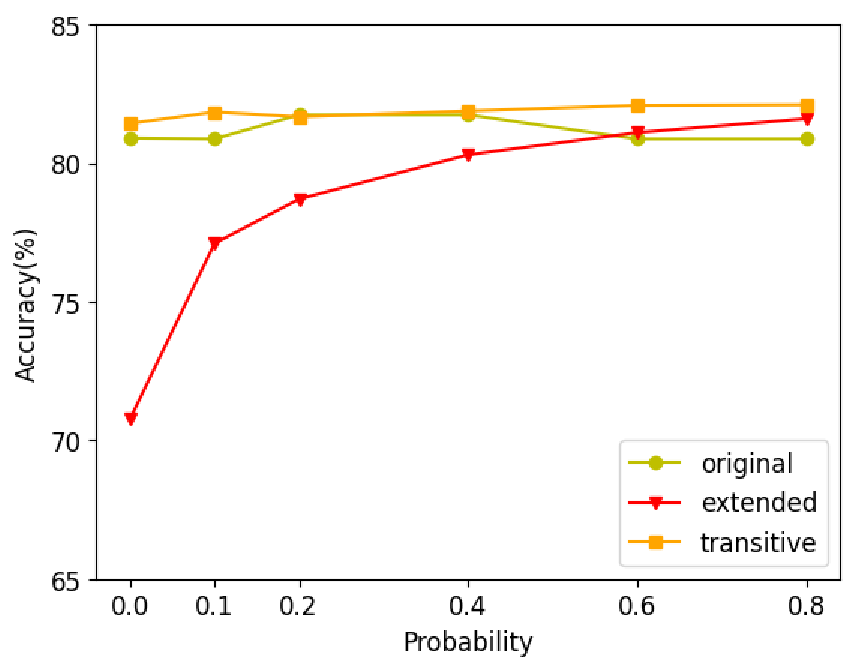}
		\caption{Effect of prompt insertion probability.}	% 标题
		\label{fig-prompt-prob}
	\end{minipage}	
	\begin{minipage}[t]{0.5\textwidth}	%并排放两张图片，每张占页面的0.5
	\centering
		\includegraphics[width=\textwidth]{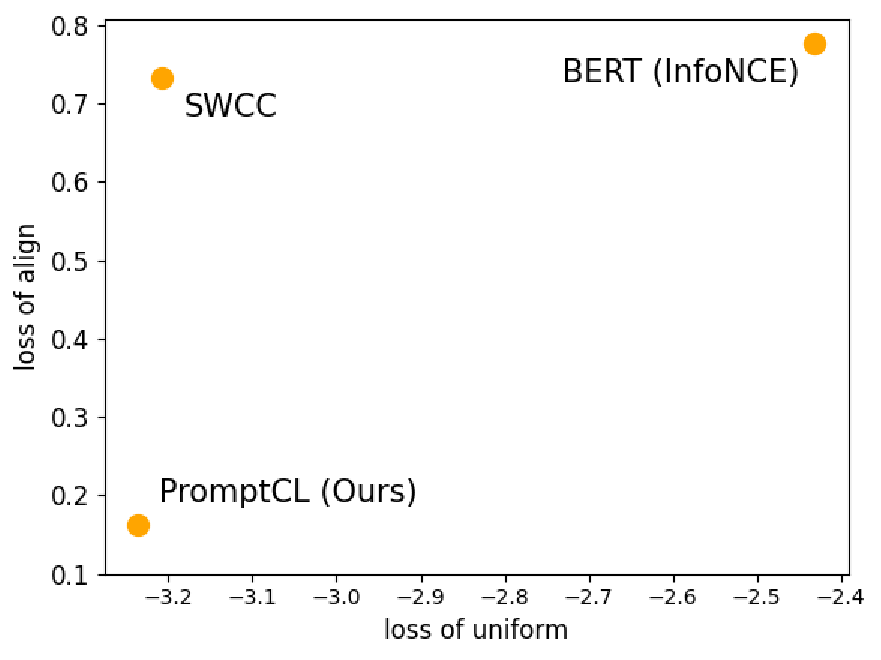}
		\caption{To plot the align loss and uniform loss. (lower is better)}	% 标题
		\label{fig-align-uniform}
	\end{minipage}
\end{figure}

\textbf{Probability of prompt insertion.}
%介绍
The influence of the probability of inserting prompt templates during the training process is depicted in Fig~\ref{fig-prompt-prob}.
%观察现象：随着提示模版插入频率的增加，模型在内部评价上的综合性能稳步增加
We have observed that as the probability of prompt template insertions increases, the model's overall performance in intrinsic evaluation steadily improves.
%建立理论：训练事件表示模型时，插入提示模版可以提高事件表示在内部评价任务上的泛化性
The insertion of prompt templates during the training of event representation models enhances the generalization of event representation in intrinsic evaluation tasks.

\textbf{Uniformity and alignment.}
%介绍
Figure 3 displays the uniformity and alignment of various event representation models, along with their Transitive Sentence Similarity results.
%观察现象：对齐性和均匀性越好，在Trans任务上的表现就越好，这与前人工作的发现是一致的
In general, models that exhibit better alignment and uniformity achieve superior performance, which confirms the findings in Wang, et al.~\cite{wang-etal-2020-understanding}.
%建立理论：相较于基线系统，在事件表示学习的过程中插入提示模版，可以显著地提升对齐性和均匀性
Additionally, we observe that the insertion of prompt templates during event representation learning significantly improves alignment compared to baselines.

\begin{table}[h]
\caption{
To demonstrate the semantic clarity of prompts and evaluate their performance.
}
\label{table-prompt}
\centering
\begin{tabular}{|l|l|l|l|l|}
\hline
prompt                            					& Original$(\%)$		& Extended$(\%)$			& Transitive$(\rho)$	\\
\hline
$x_s$ $x_p$ $x_o$         							& 80.0     			& 70.8     					& 0.81					\\
subject $x_s$ predicate $x_p$ object $x_o$         	& 79.1     			& 72.4     					& 0.82					\\
subject : $x_s$ predicate : $x_p$ object : $x_o$    & \textbf{81.7} 		& 76.7						& 0.82					\\
subject is $x_s$ predicate is $x_p$ object is $x_o$	& \textbf{81.7} 		& \textbf{78.7} 			& \textbf{0.82}			\\
\hline
\end{tabular}
\end{table}

\textbf{Content of prompt.} 
%介绍
Table~\ref{table-prompt} illustrates the impact of adjusting prompt content on the training process.
%观察现象：提示模版的语义越清晰，事件表示在Hard Similarity Task上的性能越好
As shown in the table, an increase in prompt semantic clarity results in a better performance on the Hard Similarity Tasks.
%建立理论：训练事件表示模型时，使用的提示模版语义越清晰，得到的事件表示模型在评价任务上的泛化性越好
The generalization of event representation models is closely related to the clarity of the prompt template used during training. 
Specifically, a clearer prompt template provides enhanced semantic guidance, leading to more effective event representation models with better generalization capabilities.

\begin{table}[h]
\caption{
	A case study on the Extended dataset of Hard Similarity Task.
}
\label{table-case-study}
\centering
\begin{tabular}{|l|l|l|l|l|}
\hline
Event A               & Event B                      & \begin{tabular}[c]{@{}l@{}}BERT\\ (InfoNCE)\end{tabular} & \begin{tabular}[c]{@{}l@{}}PromptCL\\ (Ours)\end{tabular} & Label \\
\hline
we focus on issues    & he pay attention to problems & 0.46                                                     & 0.61                                                      & 1     \\
we focus on issues    & we focus on people           & 0.72                                                     & 0.57                                                      & 0     \\
\hdashline
he flee city          & i leave town                 & 0.62                                                     & 0.72                                                      & 1     \\
he flee city          & he flee hotel                & 0.68                                                     & 0.41                                                      & 0     \\
\hdashline
he explain things     & she analyze problems         & 0.44                                                     & 0.50                                                      & 1     \\
he explain things     & he explain love              & 0.51                                                     & 0.31                                                      & 0     \\
%\hdashline
%i experience problems & she undergo trouble          & 0.72                                                     & 0.78                                                      & 1     \\
%i experience problems & i experience symptoms        & 0.79                                                     & 0.69                                                      & 0     \\
%\hdashline
%i obtain idea         & we propose thought           & 0.67                                                     & 0.75                                                      & 1     \\
%i obtain idea         & i obtain implementation      & 0.68                                                     & 0.66                                                      & 0     \\
%\hdashline
%he examine situation  & they investigate circumstance & 0.65                                                     & 0.78                                                      & 1     \\
%he examine situation  & he examine system             & 0.65                                                     & 0.58                                                      & 0     \\
\hline
\end{tabular}
\end{table}

\textbf{Case study.}
Table~\ref{table-case-study} shows the case study of randomly sampling several groups of events from the Extended dataset of the Hard Similarity Task. 
The performance of BERT(InfoNCE) and PromptCL in predicting the similarity scores of these events was evaluated. 
A closer alignment between the predicted and ground truth similarity scores indicates a deeper understanding of the event by the model. 
%The results of this case study are presented in Table~\ref{table-case-study}, which demonstrate that PromptCL outperforms BERT(InfoNCE) in predicting similarity scores that more closely align with the ground truth labels. 
The results are presented in Table~\ref{table-case-study}, which demonstrate that PromptCL outperforms BERT(InfoNCE) in predicting similarity scores that more closely align with the ground truth labels. 
This suggests that the proposed prompt template-based contrastive learning, and SPO word order, and EventMLM can aid in comprehending short event texts and provide semantic guidance for PLMs.

\section{Conclusion}

%In this work, we propose a simple and effective framework (SWCC) that learns event representations by making better use of co-occurrence information of events, without any addition annotations. 
This study presents a novel framework called PromptCL, which aims to improve the learning of event representations through the use of PLMs, without the need for additional features such as co-occurrence information of events as used in SWCC.
%In particular, we introduce a weakly supervised contrastive learning method that allows us to consider multiple positives and multiple negatives, and a prototype-based clustering method that avoids semantically related events being pulled apart. 
In particular, we introduce a prompt template-based contrastive learning method and SPO word order that allow us to easily elicit the text comprehension ability of PLMs, and an EventMLM method that trains the PLM to comprehend the relationships between event components.
%Our experiments indicate that our approach not only outperforms other baselines on several event related tasks, but has a good clustering performance on events. 
Our experiments demonstrate that PromptCL achieves superior performance compared to state-of-the-art baselines on several event-related tasks.
%We also provide a thorough analysis of the prototype-based clustering method to demonstrate that the learned prototype vectors are able to implicitly capture various relations between events.
Moreover, our comprehensive analysis reveals that the utilization of a prompt leads to enhanced generalization capabilities for event representations.

\subsection{Acknowledgments}
This work is supported by grants from the National Natural Science Foundation of China (No. 62076048), the Science and Technology Innovation Foundation of Dalian (2020JJ26GX035).

\bibliographystyle{splncs04}
\bibliography{nlpcc2023}

\end{document}